%% file: cvpr_Nov.tex
\documentclass[10pt,twocolumn,letterpaper]{article}

\usepackage{cvpr}
\usepackage{times}
\usepackage{epsfig}
\usepackage{graphicx}
\usepackage{amsmath}
\usepackage{amssymb}

\include{macrofile}

\usepackage{graphics}
\usepackage{algorithmic,algorithm}
\usepackage{amsthm}                 
\usepackage[mathscr]{eucal}         
\usepackage{amsbsy}                 
\usepackage{bm}                     
\usepackage{paralist}               
\usepackage{xspace}                 

\newcommand{\barr}{\left[ \begin{array} }
\newcommand{\earr}{ \end{array} \right] }
\newcommand{\ars}[1]{\left[ \begin{array}{#1}}
\newcommand{\are}{\end{array} \right] }
\newcommand{\oars}[1]{\begin{array}{#1}}
\newcommand{\oare}{\end{array}}
\newcommand{\eqs}{\begin{eqnarray}}
\newcommand{\eqe}{\end{eqnarray}}
\newcommand{\eqsn}{\begin{eqnarray*}}
\newcommand{\eqen}{\end{eqnarray*}}

\newcommand{\ens}{\begin{enumerate}}
\newcommand{\ene}{\end{enumerate}}
\newcommand{\its}{\begin{itemize}}
\newcommand{\ite}{\end{itemize}}
\newcommand{\des}{\begin{description}}
\newcommand{\dee}{\end{description}}

\newtheorem{theorem}{Theorem}[section]
\numberwithin{theorem}{subsection}
\newtheorem{lemma}{Lemma}[section]
\numberwithin{lemma}{subsection}
\newtheorem{defn}{\indent \bf Definition}[section]
\numberwithin{defn}{subsection}

\numberwithin{conjecture}{subsection}

\numberwithin{remark}{subsection}

\usepackage[breaklinks=true,bookmarks=false]{hyperref}
 \cvprfinalcopy 

\def\cvprPaperID{****} 

\ifcvprfinal\fi
\begin{document}

\title{Novel methods for multilinear data completion and de-noising based on tensor-SVD}

\author{Zemin Zhang, Gregory Ely, Shuchin Aeron\\
Department of ECE, Tufts University\\
Medford, MA 02155\\
{\tt\small zemin.zhang@tufts.com}\\
{\tt\small gregoryely@gmail.com}\\
{\tt\small shuchin@ece.tufts.edu}\\
\and
Ning Hao and Misha Kilmer\\
Department of Mathematics\\
Medford, MA 02155\\
{\tt\small ning.hao@tufts.edu}\\
{\tt\small misha.kilmer@tufts.edu}\\
}

\maketitle
\thispagestyle{empty}

\begin{abstract}
In this paper we propose novel methods for completion (from limited samples) and de-noising of multilinear (tensor) data and as an application consider 3-D and 4-D (color) video data completion and de-noising. We exploit the recently proposed tensor-Singular Value Decomposition (t-SVD)\cite{KilmerBramanHooverHao_SIAM}. Based on t-SVD, the notion of multilinear rank and a related tensor nuclear norm was proposed in \cite{KilmerBramanHooverHao_SIAM} to characterize informational and structural complexity of multilinear data.  We first show that videos with linear camera motion can be represented more efficiently using t-SVD compared to the approaches based on vectorizing or flattening of the tensors. Since efficiency in representation implies efficiency in recovery, we outline a tensor nuclear norm penalized algorithm for video completion from missing entries. Application of the proposed algorithm for video recovery from missing entries is shown to yield a superior performance over existing methods. We also consider the problem of tensor robust Principal Component Analysis (PCA) for de-noising 3-D video data from sparse random corruptions. We show superior performance of our method compared to the matrix robust PCA adapted to this setting as proposed in \cite{Candes:2011}.
\end{abstract}

\input{Introduction}
\input{tSVD}
\input{compress}

\input{completion}

\input{rPCA}
\input{Conclusion}

{\small
\bibliographystyle{ieee}
\bibliography{bibtensor,SAbib,elmbib,ZZbib}
}

\end{document}

%% file: macrofile.tex
\newcommand{\Sec}[1]{\hyperref[sec:#1]{\S\ref*{sec:#1}}} 
\newcommand{\Eqn}[1]{\hyperref[eq:#1]{(\ref*{eq:#1})}} 
\newcommand{\Fig}[1]{\hyperref[fig:#1]{Figure~\ref*{fig:#1}}} 
\newcommand{\Tab}[1]{\hyperref[tab:#1]{Table~\ref*{tab:#1}}} 
\newcommand{\Thm}[1]{\hyperref[thm:#1]{Theorem~\ref*{thm:#1}}} 
\newcommand{\Lem}[1]{\hyperref[lem:#1]{Lemma~\ref*{lem:#1}}} 
\newcommand{\Prop}[1]{\hyperref[prop:#1]{Property~\ref*{prop:#1}}} 
\newcommand{\Cor}[1]{\hyperref[cor:#1]{Corollary~\ref*{cor:#1}}} 
\newcommand{\Def}[1]{\hyperref[def:#1]{Definition~\ref*{def:#1}}} 
\newcommand{\Alg}[1]{\hyperref[alg:#1]{Algorithm~\ref*{alg:#1}}} 
\newcommand{\Ex}[1]{\hyperref[ex:#1]{Example~\ref*{ex:#1}}} 

\newcommand{\Real}{\mathbb{R}}

\newcommand{\Tra}{^{\rm T}} 

\newcommand{\M}[1]{{\bm{\mathbf{\MakeUppercase{#1}}}}} 

\newcommand{\T}[1]{\boldsymbol{\mathscr{\MakeUppercase{#1}}}} 

\newcommand{\TV}{\T{V}}



%% file: Introduction.tex
\section{Introduction}
\vspace{-2mm} 
This paper focuses on several novel methods for robust recovery of multilinear signals or tensors (essentially viewed as 2-D, 3-D,..., N-D data) under limited sampling and measurements. Signal recovery from partial measurements, sometimes also referred to as the problem of data completion for specific choice of measurement operator being a simple downsampling operation, has been an important area of research, not only for statistical signal processing problems related to inversion, \cite{Ely_ICASSP2013, Wright12-II,Ji_PAMI12}, but also in machine learning for online prediction of ratings, \cite{Hazan_JMLR12}. All of these applications exploit low structural and informational complexity of the data, expressed either as low rank for the 2-D matrices \cite{Ely_ICASSP2013,Wright12-II}, which can be extended to higher order data via \emph{flattening or vectorizing} of the tensor data such as tensor N-rank \cite{GandyRY2011}, or other more general tensor-rank measures based on particular tensor decompositions such as higher oder SVD (HOSVD) or Tucker-3 and Canonical Decomposition (CANDECOMP). See \cite{SIREV} for a survey of these decompositions.

The key idea behind these methods is that under the assumption of low-rank of the underlying data thereby constraining the complexity of the hypothesis space, it should be feasible to recover data (or equivalently predict the missing entries) from number of measurements in proportion to the rank. Such analysis and the corresponding identifiability results are obtained by considering an appropriate complexity penalized recovery algorithm under observation constraints, where the measure of complexity, related to the notion of rank, comes from a particular \emph{factorization} of the data. Such algorithms are inherently combinatorial and to alleviate this difficulty one looks for the tightest convex relaxations of the complexity measure, following which the well developed machinery of convex optimization as well as convex analysis can be employed to study the related problem. For example, rank of the 2-D matrix being relaxed to the Schatten 1-norm, \cite{Watson92} and tensor $N$-rank for order $N > 2$ tensors being relaxed to overlapped Schatten p-norms, \cite{GandyRY2011}. 

Note that all of the current approaches to handle multilinear data extend the nearly optimal 2-D SVD\footnote{Optimality of 2-D SVD is based on the optimality of truncated SVD as the best $k$-dimensional $\ell_2$ approximation.} based vector space approach to the higher order ($N > 2$) case. This results in loss of optimality in the representation.  In contrast, our approach is based upon recent results on decomposition/factorization of tensors in \cite{Braman2010,KilmerMartin2010,KilmerBramanHooverHao_SIAM} in which the authors refer to as tensor-SVD or t-SVD for short. Essentially the t-SVD is based on an operator theoretic interpretation of third-order tensors as linear operators on the space of oriented matrices \cite{Braman2010}. This notion can be extended recursively to higher order tensors \cite{Martin_SIAM2013}. In this paper we will exploit this decomposition, the associated notion of tensor multi-rank and its convex relaxation to the corresponding Tensor Nuclear Norm (TNN) (see\cite{semerci1}) for completion and recovery of multilinear data.


This paper is organized as follows. Section \ref{sec:tSVD} presents the notations and provide an overview and key results on t-SVD from \cite{Braman2010, KilmerMartin2010, KilmerBramanHooverHao_SIAM} and illustrates the key differences and advantages over other tensor decomposition methods. We will then provide an over-view of the related structural complexity measures. In  Section \ref{sec:compression} we study the compression performance of the t-SVD based representation on several video data sets. Following that, in Section \ref{sec:tcompletion} we propose a tensor nuclear norm (TNN) penalized algorithm for 3-D and 4-D (color) video completion from randomly sampled data cube. In Section \ref{sec:rPCA} we consider a tensor robust Principal Component Analysis (PCA) problem for videos with sparse data corruption and propose an algorithm to separate low multi-rank video from sparse corruptions. Finally we conclude in Section \ref{sec:future_work}.

%% file: tSVD.tex
\section{Brief overview of t-SVD}
\label{sec:tSVD}
\vspace{-2mm} 
In this section, we describe the tensor decomposition as proposed in \cite{Braman2010,KilmerMartin2010,KilmerBramanHooverHao_SIAM} and the notations used throughout the paper. 
\vspace{-2mm} 
\subsection{Notation and Indexing}
\vspace{-2mm} 
A \emph{\textbf{Slice}} of an N-dimensional tensor is a 2-D section defined by fixing all but two indices. A \emph{\textbf{Fiber}} of an N-dimensional tensor is a 1-D section defined by fixing all indices but one \cite{SIREV}. For a third order tensor $\T{A}$, we will use the Matlab notation $\T{A}(k,:,:)$ , $\T{A}(:,k,:)$ and $\T{A}(:,:,k)$ to denote the $k_{th}$ horizontal, lateral and frontal slices, and $\T{A}(:,i,j)$, $\T{A}(i,:,j)$ and $\T{A}(i,j,:)$ to denote the $(i,j)_{th}$ mode-1, mode-2 and mode-3 fiber.  In particular, we use $\T{A}^{(k)}$ to represent $\T{A}(:,:,k)$.

One can view a 3-D tensor of size $n_1 \times n_2 \times n_3$ as an $n_1 \times n_2$ matrix of tubes. By introducing a commutative operation $*$ between the tubes $\mathbf{a}, \mathbf{b}\in \Real^{1 \times 1 \times n_3}$  via $\mathbf{a} * \mathbf{b} = \mathbf{a} \circ \mathbf{b}$, where $\circ$ denotes the \emph{circular convolution} between the two vectors, one defines the t-product between two tensors as follows. 
\begin{defn} \emph{\textbf{t-product}}.  
The t-product $\T{C}=\T{A}*\T{B}$ of $\T{A} \in \mathbb{R}^{n_1 \times n_2 \times n_3}$ and $\T{B} \in \mathbb{R}^{n_2\times n_4 \times n_3}$ is a tensor of size $n_1 \times n_4 \times n_3$ where the $(i,j)_{th}$ tube denoted by $\T{C}(i,j,:)$ for $i = 1,2, ..., n_1$ and $j = 1, 2,..., n_4$ of the tensor $\T{C}$ is given by $\sum_{k=1}^{n_2} \T{A}(i,k,:) * \T{B}(k,j,:)$.
\end{defn}
The t-product is analogous to the matrix multiplication except that circular convolution replaces the multiplication operation between the elements, which are now tubes. Next we define related notions of tensor transpose and identity tensor.
\begin{defn}
\emph{\textbf{Tensor Transpose}}.  Let $\T{A}$ be a tensor of size $n_1 \times n_2 \times n_3$, then $\T{A} \Tra$ is the $n_2 \times n_1 \times n_3$ tensor obtained by transposing each of the frontal slices and then reversing the order of transposed frontal slices $2$ through $n_3$.
\end{defn}
\begin{defn} \emph{\textbf{Identity Tensor}}. The identity tensor $\T{I} \in \mathbb{R}^{n_1 \times n_1 \times n_3}$ is a tensor whose first frontal slice is the $n_1 \times n_1$ identity matrix and  all other frontal slices are zero.
\end{defn}
\begin{defn} \emph{\textbf{f-diagonal Tensor}}. A tensor is called f-diagonal if each frontal slice of the tensor is a diagonal matrix.
\end{defn}

The t-product of $\T{A}$ and $\T{B}$ can be computed efficiently by performing the fast Fourier transformation (FFT) along the tube fibers of $\T{A}$ and $\T{B}$ to get $\hat{\T{A}}$ and $\hat{\T{B}}$,  multiplying the each pair of the frontal slices of $\hat{\T{A}}$ and $\hat{\T{B}}$ to obtain $\hat{\T{C}}$, and then taking the inverse FFT along the third mode to get the result. For details about the computation, see \cite{KilmerMartin2010, KilmerBramanHooverHao_SIAM}.
\begin{defn} \emph{\textbf{Orthogonal Tensor}}. A tensor $\T{Q}\in\mathbb{R}^{n_1\times n_1\times n_3}$ is orthogonal if
\begin{equation}
\T{Q} \Tra * \T{Q} = \T{Q} * \T{Q} \Tra  = \T{I}
\end{equation}
where $*$ is the t-product.
\end{defn}

\subsection{Tensor Singular Value Decomposition (t-SVD)}
\begin{figure}[htbp]
\centering \makebox[0in]{
    \begin{tabular}{c c}
      \includegraphics[scale=0.45]{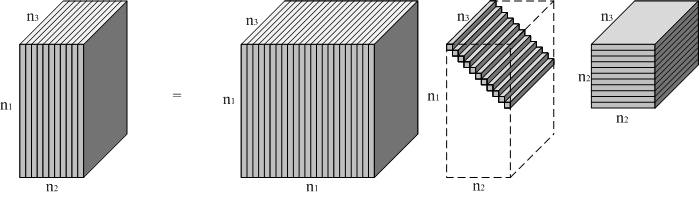}
 \end{tabular}}
  \caption{ The t-SVD of an $n_1 \times n_2 \times n_3$ tensor.}
  \label{fig:tSVD}
\end{figure}

\vspace{-2mm} 
The new t-product  allows us to define a tensor Singular Value Decomposition (t-SVD). 
\begin{theorem}
For $\T{M}\in\mathbb{R}^{n_1 \times n_2 \times n_3}$, the t-SVD of $\T{M}$ is  given by
\begin{equation}
\T{M} = \T{U} *\T {S} *\T{V}\Tra
\end{equation}
\noindent where $\T{U}$ and $\TV$ are orthogonal tensors of size $n_1 \times n_1 \times n_3 $ and  $n_2 \times n_2 \times n_3 $ respectively. $\T{S}$ is a rectangular $f$-diagonal tensor of size $n_1 \times n_2 \times n_3 $, and $*$ denotes the t-product. 
\end{theorem}
\vspace{-1.5mm}
One can obtain this decomposition by computing matrix SVDs in the Fourier domain, see Algorithm 1. The notation in the algorithm can be found in \cite{Martin_SIAM2013}. 
Figure \ref{fig:tSVD} illustrates the decomposition for the 3-D case.
\begin{algorithm}
\label{alg:tSVD - for $N$-D tensor}
  \caption{t-SVD}
  \begin{algorithmic}
  \STATE \textbf{Input: } $\T{M} \in \mathbb{R}^{n_1 \times n_2  ... \times n_N}$
  \STATE $\rho = n_3n_4...n_N$
   \FOR{$i = 3 \hspace{2mm} \rm{to} \hspace{2mm} N$}
  	\STATE ${\T{D}} \leftarrow \rm{fft}(\T{M},[\hspace{1mm}],i)$;
  \ENDFOR
  
  \FOR{$i = 1 \hspace{2mm} \rm{to} \hspace{2mm} \rho$}
  	\STATE $ [\M{U}, \M{S}, \M{V}] = SVD(\T{D}(:,:,i))$
  	\STATE $ {\hat{\T{U}}}(:,:,i) = \M{U}; \hspace{1mm} {\hat{\T{S}}}(:,:,i) = \M{S}; \hspace{1mm} {\hat{\T{V}}}(:,:,i) = \M{V}; $
  \ENDFOR
  
  \FOR{$i = 3 \hspace{2mm} \rm{to} \hspace{2mm} N$}
  	\STATE $\T{U} \leftarrow \rm{ifft}(\hat{\T{U}},[\hspace{1mm}],i); \hspace{1mm} \T{S} \leftarrow \rm{ifft}(\hat{\T{S}},[\hspace{1mm}],i); \hspace{1mm} \T{V} \leftarrow \rm{ifft}(\hat{\T{V}},[\hspace{1mm}],)i$;
  \ENDFOR
  \end{algorithmic}
\end{algorithm}

%
%

\vspace{-2mm} 
\subsection{t-SVD: Fundamental theorems and key results}
\vspace{-2mm} 
The two widely used tensor decompositions, Tucker and PARAFAC\cite{SIREV} are usually seen as a higher order SVD for tensors. Both of these decompositions have several disadvantages. In particular, one cannot  easily determine the rank-one components of the PARAFAC decomposition and given a fixed rank, calculation of an approximation can be  numerically unstable. The tensor-train form of the Tucker decomposition is studied in \cite{Oseledets} as an alternative form. Tucker decomposition can be seen as a generalization of PARAFAC decomposition, and the truncated decomposition doesn't yield the best fit of the original tensor. In contrast, the t-SVD can be easily computed by solving several SVDs in the Fourier domain. More importantly, it gives an optimal approximation of a tensor measured by the Frobenious norm of the difference, as stated in the following theorem\cite{KilmerMartin2010, KilmerBramanHooverHao_SIAM, HaoKilmerBramanHoover2012}. 

\begin{theorem} Let the t-SVD of $\T{M} \in \mathbb{R}^{n_1 \times n_2 \times n_3}$ be 
given by $\T{M} = \T{U} * \T{S} * \T{V}^T$ and for $k < \min(n_1,n_2)$ define
$\T{M}_k = \sum_{i=1}^{k} \T{U}(:,i,:) * \T{S}(i,i,:) * \T{V}(:,i,:)^T $, 
Then
$$\T{M}_k = \arg \min_{\tilde{\T{M}} \in \mathbb{M}} \| \T{M} - \tilde{\T{M}} \|_F $$
where $\mathbb{M} = \{ \T{C} = \T{X} * \T{Y} |
\T{X} \in \mathbb{R}^{n_1 \times k \times n_3}, \T{Y} \in \mathbb{R}^{k 
\times n_2 \times n_3} \}$.
\end{theorem}
\vspace{-2mm} 
\subsection{Measures of tensor complexity using t-SVD }
\vspace{-2mm} 
We  now define two measures of tensor complexity based on the proposed t-SVD: the tensor multi-rank, proposed in  \cite{KilmerBramanHooverHao_SIAM}, and the novel tensor tubal rank.
\begin{defn} \textbf{Tensor multi-rank}. The multi-rank of $\T{A}\in\mathbb{R}^{n_1\times n_2\times n_3}$ is a vector $p$ $\in \Real^{n_3 \times 1}$ with the $i_{th}$ element equal to the rank of the $i_{th}$ frontal slice of $\hat{\T{A}}$ obtained by taking the Fourier transform along the third dimension of the tensor. 
\end{defn}
One can obtain a scalar measure of complexity as the $\ell_1$ norm of the tensor multi-rank. We now define another measure motivated by the matrix SVD. 
\begin{defn} \textbf{Tensor tubal-rank}. The tensor tubal rank of a 3-D tensor is defined to be the number of non-zero tubes of $\T{S}$ in the t-SVD factorization.
\end{defn}
As in the matrix case,  practical applications of these complexity measures require adequate convex relaxations. To this end we have the following result for the Tensor multi-rank. 

\begin{theorem}
The tensor-nuclear-norm (TNN) denoted by $||\T{A}||_{TNN}$ and defined as the sum of the singular values of all the frontal slices of $\hat{\T{A}}$ is a norm and is the tightest convex relaxation to $\ell_1$ norm of the tensor multi-rank.
\end{theorem}
\begin{proof}
\vspace{-3mm}
The proof that TNN is a valid norm can be found in \cite{semerci1}. The $\ell_1$ norm of the tensor multi-rank is equal to $\text{rank}(\text{blkdiag}(\hat{\T{A}}))$,
for which the tightest convex relaxation is the the nuclear norm of  $\text{blkdiag}(\hat{\T{A}})$ which is TNN of $\T{A}$ by definition. Here $\text{blkdiag}(\hat{\T{A}})$ is a block diagonal matrix defined as follows:
\begin{equation}
\label{eq:blkdiag}
\text{blkdiag}( \hat{\T{A}} ) = 
  \left[\begin{array}{cccc}\hat{{\T{A}}}^{(1)}& & & \\
 & \hat{{\T{A}}}^{(2)} & & \\
 & &\ddots & \\
  & & & \hat{{\T{A}}}^{(n_3)}\end{array} \right]
\end{equation}
where $\hat{\T{A}}^{(i)}$ is the $i_{th}$ frontal slice of $\hat{\T{A}}$, $i = 1,2,...,n_3$.
\end{proof}

Unlike the TNN is a relaxation for the tensor-nuclear-norm, there is no clear convex relaxation for the tensor tubal-rank. 
In the next section we will undertake a compressibility study for tensor data using two types of truncation strategies based on t-SVD and compare them with matrix SVD based approach on several video data sets.

%% file: compress.tex
\section{Multilinear data compression using t-SVD}
\label{sec:compression}
We outline two methods for compression based on t-SVD and compare them with the traditional truncated SVD based approach in this section. Note that we don't compare with truncated HOSVD or other tensor decompositions as there is no notion of optimality for these decompositions in contrast to truncated t-SVD and truncated SVD. 

The use of SVD in matrix compression has been widely studied in \cite{465543}. 
For a matrix $A \in \Real^{m \times n}$ with its SVD $A = USV^T$, the rank $r$ approximation of $A$ is the matrix $A_r = U_r S_r V^\text{T}_r$, where $S_r$ is a $r \times r$ diagonal matrix with $S_r(i,i) = S(i,i), i = 1,2,...,r$. $U_r$ consists of the first $r$ columns of $U$ and $V^\text{T}_r$ consists of the first $r$ rows of $V^\text{T}_r$. The compression is measured by the ratio of the total number of entries in $A$, which is $mn$, to the total number of entries in $U_r$, $S_r$ and $V^\text{T}_r$, which is equal to $(m+n+1)r$. Extending this approach to a third-order tensor $\T{M}$ of size $n_1 \times n_2 \times n_3$, we vectorize each frontal slice and save it as a column, so we get an $n_1 n_2 \times n_3$ matrix. Then the compression ratio of rank $k_1$ SVD approximation is 
\begin{align}
\text{ratio}_{\text{SVD}} = \frac{n_1 n_2 n_3}{n_1 n_2 k_1 + k_1 + n_3 k_1} = \frac{n_1 n_2 n_3}{k_1(n_1 n_2 +n_3+1)}
\end{align}
where $1 \le k_1 \le \text{min}(n_1n_2,n_3)$. Generally even with small $k_1$, the approximation $M_{k_1}$ gets most of the information of $\T{M}$. 

{\bf Method 1} : Based on t-SVD our first method for compression, which we call t-SVD compression, basically follows the same idea of truncated SVD but in the \emph{Fourier domain}. For an $n_1 \times n_2 \times n_3$ tensor $\T{M}$, we use Algorithm 1 to get $\hat{\T{M}}$, $\hat{\T{U}}$, $\hat{\T{S}}$ and $\hat{\T{V}}^\text{T}$. It is known that $\hat{\T{S}}$ is a f-diagonal tensor with each frontal slice is a diagonal matrix. So the total number of f-diagonal entries of $\hat{\T{S}}$ is $n_0 n_3$ where $n_0 = \text{min} (n_1,n_2)$.  We choose an integer $k_2$, $1 \le k_2 \le n_0 n_3$ and keep the $k_2$ largest f-diagonal entries of $\hat{\T{S}}$ then set the rest to be $0$. If $\hat{\T{S}}(i,i,j)$ is set to be $0$, then let the corresponding columns $\hat{\T{U}}(:,i,j)$ and $\hat{\T{V}}^\text{T}(:,i,j)$ also be $0$. We then call the resulting tensors $\hat{\T{U}}_{k_2}$, $\hat{\T{S}}_{k_2}$ and $\hat{\T{V}}^\text{T}_{k_2}$. So the approximation is $\T{M}_{k_2} = \T{U}_{k_2} * \T{S}_{k_2} * \T{V}^\text{T}_{k_2}$ where $\T{U}_{k_2}$, $\T{S}_{k_2}$ and $\T{V^\text{T}}_{k_2}$ are the inverse Fourier transforms of $\hat{\T{U}}_{k_2}$, $\hat{\T{S}}_{k_2}$ and $\hat{\T{V}}^\text{T}_{k_2}$  along the third dimension.  
The compression ratio rate for this method is
\begin{align}
\nonumber \text{ratio}_\text{t-SVD} 
& = \frac{n_1 n_2 n_3}{k_2(n_1+n_2+1)}
\end{align}
where $1 \le k_2 \le n_0 n_3$.

{\bf Method 2}:  Our second method for compressing is called t-SVD-tubal compression and is also similar to truncated SVD but in the \emph{t-product} domain. As in \textbf{Theorem 2.3.1}, we take the first $k_3$ tubes $ ( 1\le k_3 \le n_0) $ in $\T{S}$ and get the approximation $\T{M}_{k_3} = \sum_{i=1}^{k_3} \T{U}(:,i,:) * \T{S}(i,i,:) * \T{V}(:,i,:)^\text{T} $.

Compression ratio rate for the second method is 
\begin{align}
\nonumber \text{ratio}_\text{t-SVD-tubal} 
&= \frac{n_1 n_2}{k_3(n_1 + n_2 +1)}
\end{align}
where $1 \le k_3 \le n_0$.

%
%
%

\begin{figure}[h]
\centering \makebox[0in]{
    \begin{tabular}{c c}
    \includegraphics[width = .16\textwidth]{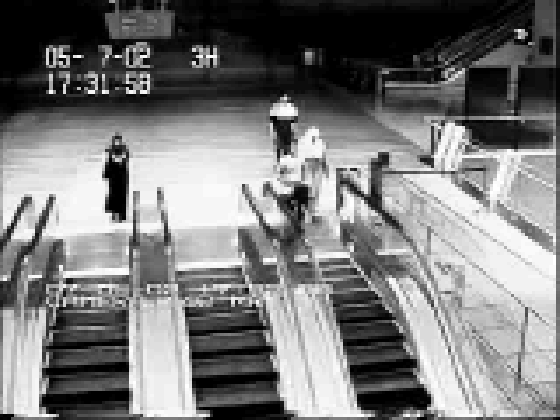}
      \includegraphics[width = .16\textwidth]{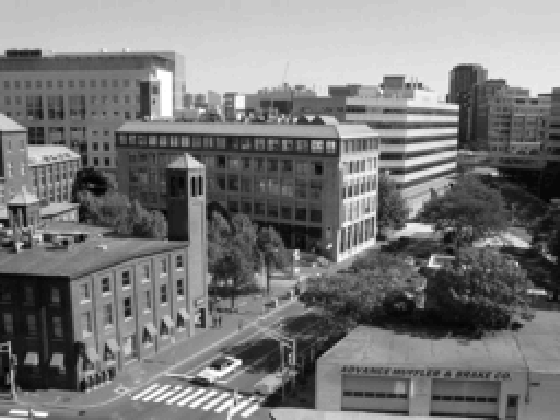}
      \includegraphics[width = .16\textwidth]{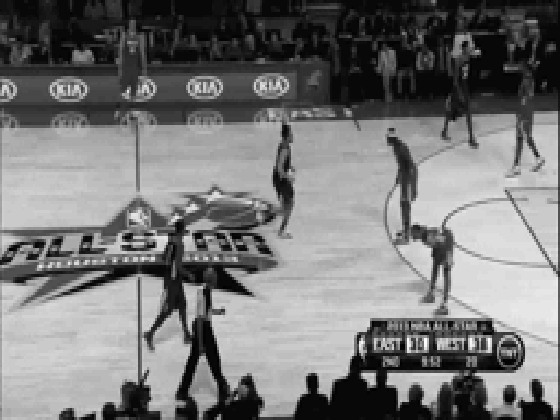}\\
      (a)\\
      \includegraphics[width = .16\textwidth]{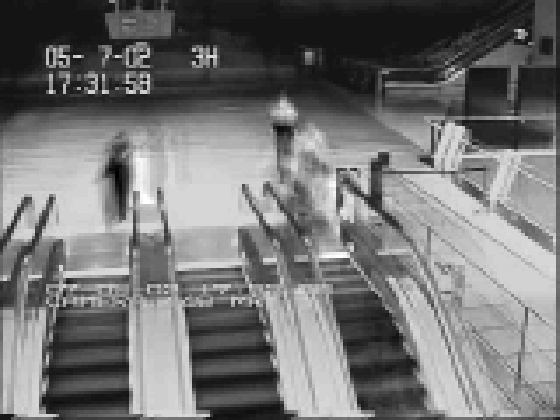}  
      \includegraphics[width = .16\textwidth]{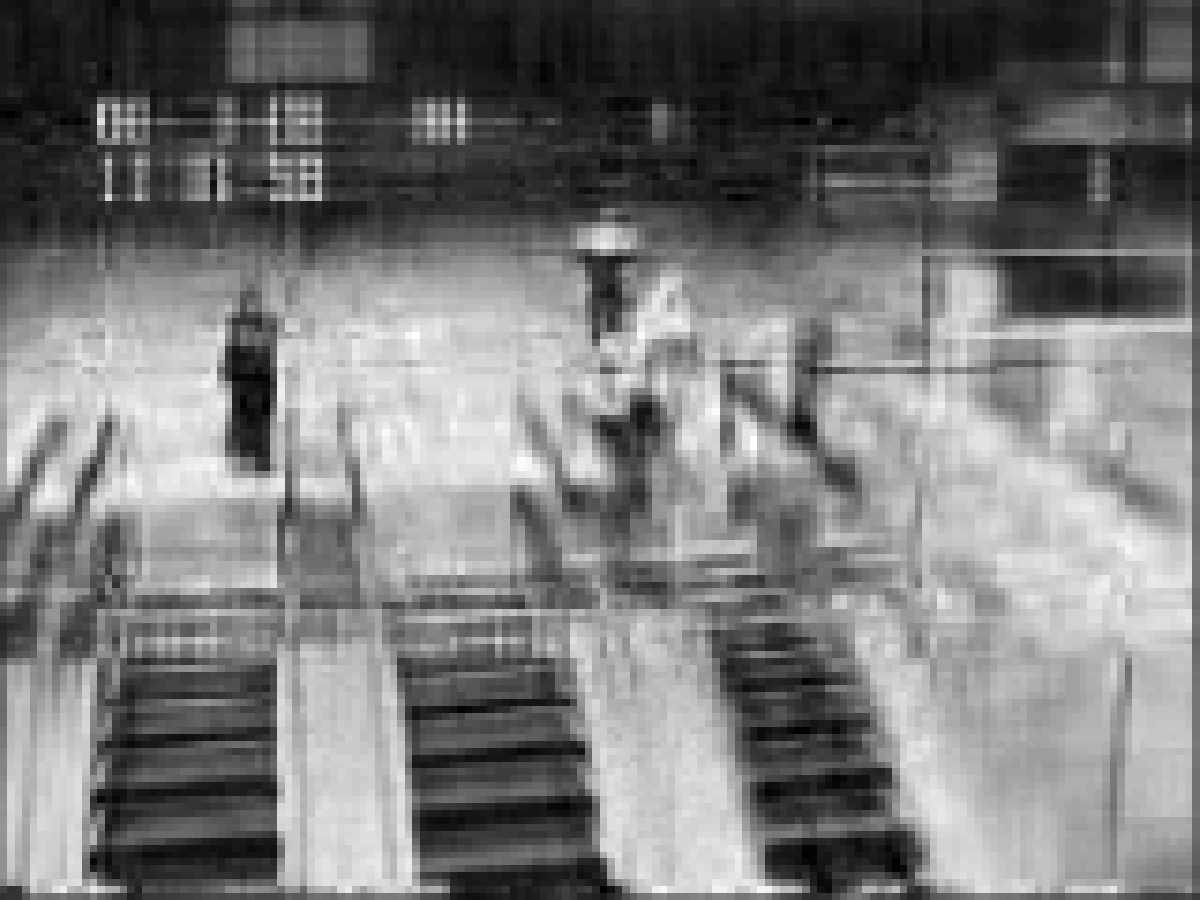} 
      \includegraphics[width = .16\textwidth]{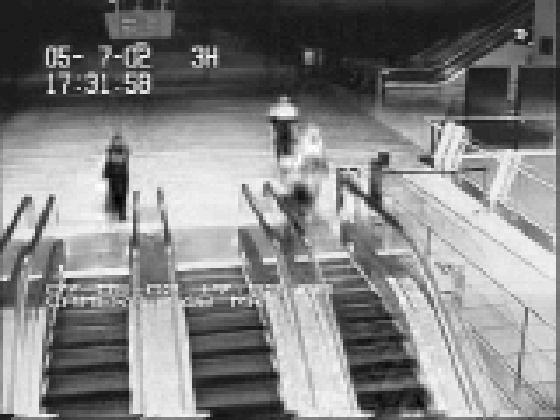} \\
      (b) \\
      \includegraphics[width = .16\textwidth]{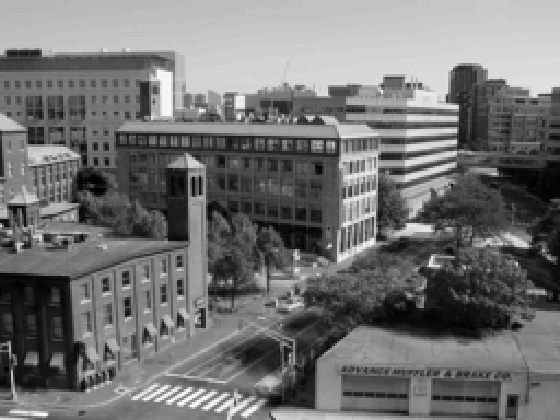}
      \includegraphics[width = .16\textwidth]{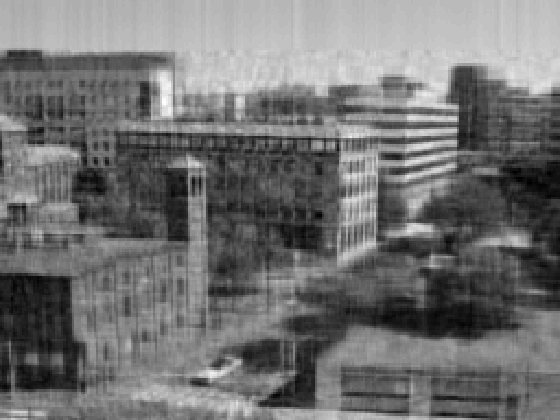}
      \includegraphics[width = .16\textwidth]{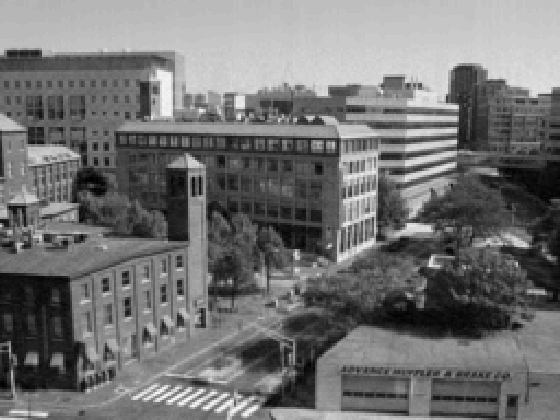}\\
      (c) \\
      \includegraphics[width = .16\textwidth]{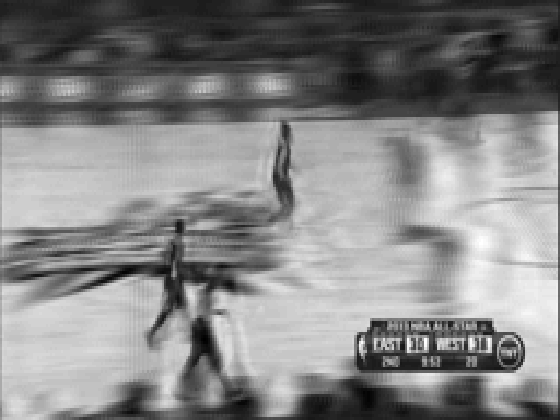}
      \includegraphics[width = .16\textwidth]{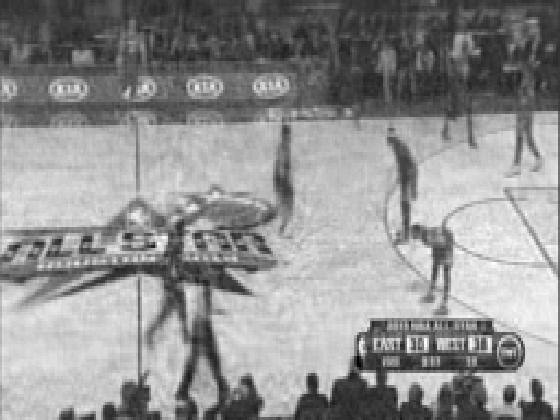}
      \includegraphics[width = .16\textwidth]{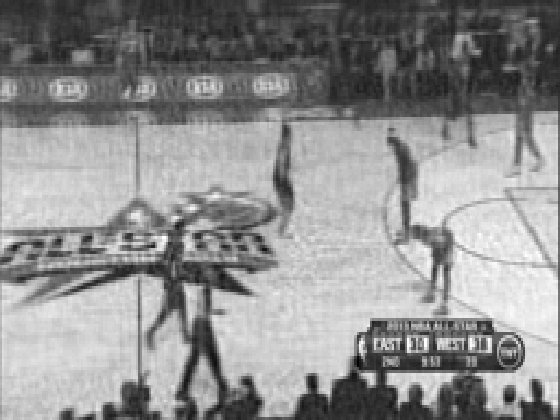} \\
      (d) 
      \end{tabular}}
  \caption{ (\textbf{a}) Three testing videos: escalator video, MERL video and basketball video. (b) (c) (d) are compression results under compression ratio 5. For (b) (c) (d) from left to right: SVD compression, t-SVD-tubal compression and t-SVD compression}
\label{fig:compress_recovery}
\end{figure}
\begin{figure}[htbp]
\centering \makebox[0in]{
    \begin{tabular}{c c c}
      \includegraphics[width = .18\textwidth, height = 1.2in]{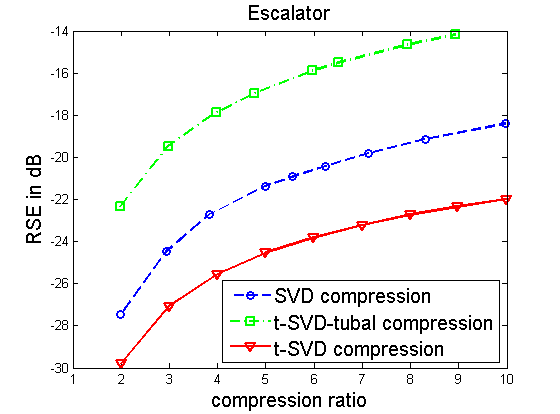}
      
      \includegraphics[width = .18\textwidth, height = 1.2in]{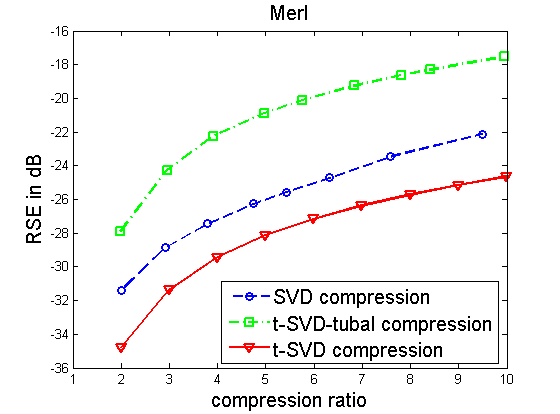}
      
      \includegraphics[width = .18\textwidth, height = 1.2in]{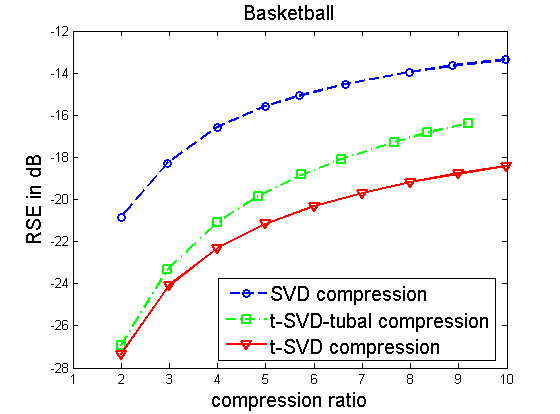}
      
      \end{tabular}}
  \caption{ Compression ratio and RSE comparison for 3 videos.}
\label{fig:compress_rse}
\end{figure}
\textbf{Video data representation and compression using t-SVD}: We now illustrate the performance of SVD based compression, t-SVD compression and t-SVD tubal compression on 3 video datasets shown in Figure~\ref{fig:compress_recovery}-(a).
\vspace{-2mm}
\begin{enumerate}
\itemsep0em
\item The first video, referred to as the Escalator video, (source: http://www.ugcs.caltech.edu/~srbecker/rpca.shtml\#2 ) of size $130 \times 160 \times 50$ (length $\times$ width $\times$ frames) from a stationary camera. 
\item The second video, referred to as the MERL video, is a {\bf \emph{time lapse video}} of size $192 \times 256 \times 38$ also from a stationary camera ({\bf data courtesy}: Dr. Amit Agrawal, Mitsubishi Electric Research Labs (MERL), Cambridge, MA). 
\item The third video, referred to as the Basketball video is a $144 \times 256 \times 80$ video (source: YouTube) with a  {\bf \emph{non-stationary panning camera}} moving from left to right horizontally following the running players. 
\end{enumerate}
\vspace{-2mm}
Figure~\ref{fig:compress_recovery} (b) to (d) show the compression results for the 3 videos when truncated according to vectorized SVD and t-SVD compression (method 1) and t-SVD tensor tubal compression (method 2). In Figure~\ref{fig:compress_rse} we show the relative square error (RSE) comparison for different compression ratio where RSE is defined in dB as $\text{RSE} = 20 \log_{10} ( \|\T{X}_{\text{rec}} - \T{X}\|_\text{F} / \|\T{X}\|_\text{F} )$. In all of the 3 results, the performance of t-SVD compression (method 1) is the best. This implies that tensor multi-rank fits very well for video datasets from both stationary and non-stationary cameras. SVD compression method (based on vectorization) has a better performance over the t-SVD-tubal compression on the Escalator and MERL video. 
However, t-SVD tubal compression (method 2) works much better than SVD compression on the Basketball video. This is because in the videos where the camera is panning or in motion, one frontal slice of the tensor to the next frontal slice can be effectively represented as a shift and scaling operation which in turn is captured by a convolution type operation and t-SVD is based on such an operation along the third dimension.

%% file: completion.tex
\vspace{-4mm}
\section{Tensor completion from limited samples}
\label{sec:tcompletion}

We will show the case when the tensor data is simply decimated randomly or down sampled in this section. Specifically we consider the problem of data completion from missing entries for multilinear signals.  Suppose there is an unknown tensor $\T{M}$ of size $n_1 \times n_2 \times n_3$ which is {\bf \emph{assumed to have a low tubal-rank}} and we are given a subset of entries $ \{ \T{M}_{ijk}:(i,j,k) \in \bold{\Omega} \}$ where $\bold{\Omega}$ is an indicator tensor of size $n_1 \times n_2 \times n_3$. Our objective is to recover the entire $\T{M}$. This section develops an algorithm for addressing this problem via solving the following complexity penalized algorithm:
\begin{equation}
\begin{aligned}
\label{eq:originalTNN}
\mbox{min} \hspace{4mm}&\|\T{X}\|_{TNN} \\
\mbox{subject to } \,\,\,&P_{\bold{\Omega}} ( \T{X} ) = P_{\bold{\Omega}} ( \T{M} )
\end{aligned}
\end{equation}
\noindent where $P_\bold{\Omega}$ is the orthogonal projector onto the span of tensors vanishing outside of $\bold{\Omega}$. So the $(i,j,k)_{th}$ component of $ P_{\bold{\Omega}} ( \T{X} )$ is equal to $\T{M}_{ijk}$ if $(i,j,k) \in \bold{\Omega}$ and zero otherwise. Let $\T{Y}$ be the available (sampled) data: $\T{Y} = P_\bold{\Omega} \T{M}$. Define $\mathcal{G}=\mathscr{F}_3 P_\bold{\Omega} \mathscr{F}_{3}^{-1}$ where $\mathscr{F}_3$ and $\mathscr{F}^{-1}_3$ are the operators representing the Fourier and inverse Fourier transform along the third dimension of tensors. Then we have
$\hat{\T{Y}} = \mathcal{G} (\hat{\T{M}})$ where $\hat{\T{Y}}$ and $\hat{\T{M}}$ are the Fourier transforms of $\T{Y}$ and $\T{M}$ along the third mode.
So (\ref{eq:originalTNN}) is equivalent with the following:
\begin{equation}
\begin{aligned}
\label{eq:fourierdomain_min}
\mbox{min} \hspace{4mm}&||\mbox{blkdiag}(\hat{\T{X}}) ||_*\\
\mbox{subject to } \,\,\,\,\,&\hat{\T{Y}} = {\cal G}(\hat{\T{X}}) 
\end{aligned}
\end{equation}
where $\hat{\T{X}}$ is the Fourier transform of $\T{X}$ along the third dimension and $\text{blkdiag}(\hat{\T{X}})$ is defined in (\ref{eq:blkdiag}). Noting that $\|\T{X}\|_{TNN} = ||\text{blkdiag}(\hat{\T{X}}) ||_*$. To solve the optimization problem, one can re-write (\ref{eq:fourierdomain_min}) equivalently as follows: 
\begin{equation}
\begin{aligned}
\min \hspace{4mm}& ||\text{blkdiag}( \hat{\T{Z}} )||_* + \mathbf{1}_{\hat{\T{Y}} = {\cal G}(\hat{\T{X}})} \\
\mbox{subject to } \,\,\,\,\, &\hat{\T{X}} - \hat{\T{Z}} = 0  
\end{aligned}
\end{equation}
where $\mathbf{1}$ denotes the indicator function. Then using the general framework of Alternating Direction Method of Multipliers(ADMM)\cite{Boyd_ADMM} we have the following recursion, 
\begin{align}
\label{eq:proj}
& \T{X}^{k+1} \nonumber \\
&= \arg\min_{\T{X}} \left\{ \mathbf{1}_{\T{Y} = P_{\Omega}(\T{X})} + \T{X}(:)\Tra \T{Q}^k(:)  + \frac{1}{2} || \T{X} - \T{Z}^{k}||_{F}^{2} \right\}\nonumber \\
& = \arg\min_{ \T{X}: \T{Y} = P_\Omega(\T{X}) } \left\{|| \T{X} - (\T{Z}^{k} - \T{Q}^{k})||_{F}^{2} \right\}\\
\label{eq:shrink}
& \hat{\T{Z}}^{k+1}  \nonumber \\
&= \arg \min_{\hat{\T{Z}}} \left\{ \frac{1}{\rho} ||\text{blkdiag}(\hat{\T{Z}})||_* + \frac{1}{2} ||  \hat{\T{Z}} - (\hat{\T{X}}^{k+1} + \hat{\T{Q}}^{k})||_{F}^{2} \right\}\\
& \T{Q}^{k+1} = \T{Q}^{k} + \left(\T{X}^{k+1} - \T{Z}^{k+1}\right)
\end{align}
where Equation~(\ref{eq:proj}) is least-squares projection onto the constraint and the solution to Equation~(\ref{eq:shrink}) is given by the singular value thresholding\cite{Watson92,Cai_SVT}. The $\T{X}(:)$ and $\T{Q}^k(:)$ means vectorizing the tensors which is Matlab notation.

\subsection{Equivalence of the algorithm to iterative singular-tubal shrinkage via convolution}
We will now show that the proposed algorithm for tensor completion has a very nice interpretation as an iterative singular tubal shrinkage using a convolution operation between the singular tubes and a tube of threshold vectors.\\

According to the particular format that (\ref{eq:shrink}) has, we can break it up into $n_3$ independent minimization problems. Let $\hat{\T{Z}}^{k+1,(i)}$ denotes the $i_{th}$ frontal slice of $\hat{\T{Z}}^{k+1}$. Similarly define $\hat{\T{X}}^{k+1,(i)}$ and $\hat{\T{Q}}^{k,(i)}$. Then (\ref{eq:shrink}) can be separated as:
\begin{equation}
\begin{aligned}
\label{eq:shrink2}
&\hat{\T{Z}}^{k+1,(i)}   \\
& = \arg \min_{W} \left\{ \frac{1}{\rho} ||W||_* + \frac{1}{2} ||  W - (\hat{\T{X}}^{k+1,(i)} + \hat{\T{Q}}^{k,(i)})||_{F}^{2} \right\}
\end{aligned}
\end{equation}
for $i = 1,2,...,n_3$. This means each $i_{th}$ frontal slice of $\hat{\T{Z}}^{k+1}$ can be calculated through (\ref{eq:shrink2}). 

In order to solve (\ref{eq:shrink2}), we give out the following lemma.
\begin{lemma}
Consider the singular value decomposition (SVD) of a matrix $X \in \mathbb{C}^{n_1 \times n_2}$ of rank $r$.

\begin{equation}
X = U \Sigma V^*,\hspace{2mm} \Sigma = \mbox{diag}(\{\sigma_i\}_{1\le i \le r}),
\end{equation}

\noindent where $U$ and $V$ are respectively $n_1 \times r$ and $n_2 \times r$ unitary matrices with orthonormal columns, and the singular values $\sigma_i$ are real and positive. Then for all $\tau \ge 0$ , define the soft-thresholding operator $D_\tau$ as follows \cite{Cai_SVT} :
\begin{equation}
\label{operator}
D_\tau(X) := U D_\tau(\Sigma) V^*, \hspace{2mm} D_\tau (\Sigma) = \mbox{diag}\{ ( \sigma_i - \tau)_+\},
\end{equation}
\noindent where $t_+$ is the positive part of t, namely, $t_+ = \mbox{max}(0,t)$.
Then, for each $\tau \ge 0$ and $Y \in C^{n_1 \times n_2}$, the singular value shrinkage operator (\ref{operator}) obeys
\begin{equation}
D_\tau(Y) = \arg\min_{X \in \mathbb{C}} \left\{ \frac{1}{2} \|X-Y\|^2_F + \tau \|X\|_* \right \}
\end{equation}
\end{lemma}

The proof can be found in \cite{Cai_SVT} for the case when the matrix is real valued. However, it can be easily extended to matrices with complex entries using the result on gradients of unitarily invariant norms in \cite{Lewis95theconvex}.

Now note that, if $U S V^\text{T} = (\hat{\T{X}}^{k+1,(i)} + \hat{\T{Q}}^{k,(i)})$ is the SVD of $(\hat{\T{X}}^{k+1,(i)} + \hat{\T{Q}}^{k,(i)})$, then the solution to (\ref{eq:shrink2}) is $U D_\tau (S) V^\text{T}$, where $D_\tau (S) = \text{diag}(S_{i,i} - \tau)_+$ for some positive constant $\tau$ and $``+"$ means keeping the positive part. This is equivalent to multiplying $(1-\frac{\tau}{S_{i,i}})_+$ to the $i_{th}$ singular value of $S$. So each frontal slice of $\hat{\T{Z}}^{k+1}$ can be calculated using this shrinkage on each frontal slice of $(\hat{\T{X}}^{k+1} + \hat{\T{Q}}^{k})$. Let  $\T{U} * \T{S}* \T{V}^\text{T} = (\T{X}^{k+1} + \T{Q}^{k})$ be the t-SVD of $(\T{X}^{k+1} + \T{Q}^{k})$ and $\hat{\T{S}}$ be the Fourier transform of $\T{S}$ along the third dimension. Then each element of the singular tubes of $\hat{\T{Z}}^{k+1}$ is the result of multiplying every entry $\hat{\T{S}}(i,i,j)$ with $(1-\frac{\tau}{\hat{\T{S}}(i,i,j)})_+$ for some $\tau > 0$. Since this process is carried out in the Fourier domain, in the original domain it is equivalent to convolving each tube $\T{S}(i,i,:)$ of $\T{S}$ with a real valued tubal vector $\vec{\tau}_i$ which is the inverse Fourier transform of the vector $[ ( 1-\frac{\tau_{i}(1)}{\hat{\T{S}}(i,i,1)})_+ , ( 1-\frac{\tau_{i}(2)}{\hat{\T{S}}(i,i,2)})_+ , ... , ( 1-\frac{\tau_{i}(n_3)}{\hat{\T{S}}(i,i,n_3)})_+ ]$. This operation can be captured by $\T{S}*\T{T}$, where $\T{T}$ is an f-diagonal tensor with $i_{th}$ diagonal tube to be $\vec{\tau}_i$. Then  $\T{Z}^{k+1} = \T{U}* ( \T{S}*\T{T} )* \T{V}^\text{T}$. In summary, the shrinkage operation in the Fourier domain on the singular values of each of the frontal faces is equivalent to performing a \emph{tubal shrinkage} via convolution in the original domain.


\textbf{Application to Video data completion} - For experiments we test 3 algorithms for video data completion from randomly missing entries:  TNN minimization of Section \ref{sec:tcompletion}, Low Rank Tensor Completion (LRTC) algorithm in \cite{Ji_PAMI12}, which uses the notion of tensor-n-rank \cite{GandyRY2011}, and the nuclear norm minimization on the vectorized video data using the algorithm in \cite{Cai_SVT}. As an application of the t-SVD to higher order tensor we also show performance on a 4-D color Basketball video data of size $144 \times 256 \times 3 \times 80$. 

\begin{figure}[htbp]
\centering \makebox[0in]{
    \begin{tabular}{c c c }
      \includegraphics[width = .17\textwidth, height = 1.2in]{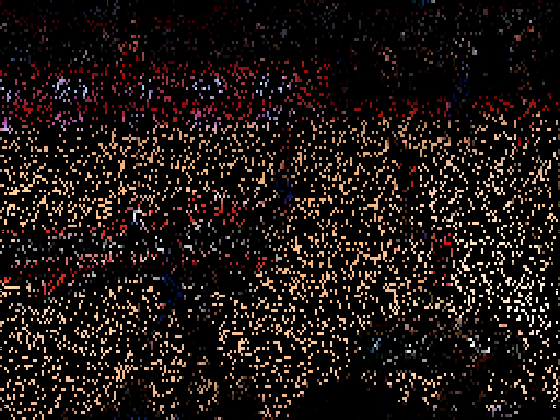}  
      \includegraphics[width = .17\textwidth, height = 1.2in]{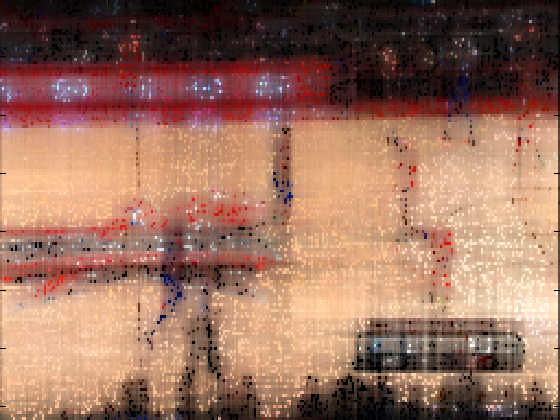}     
      \includegraphics[width = .17\textwidth, height = 1.2in]{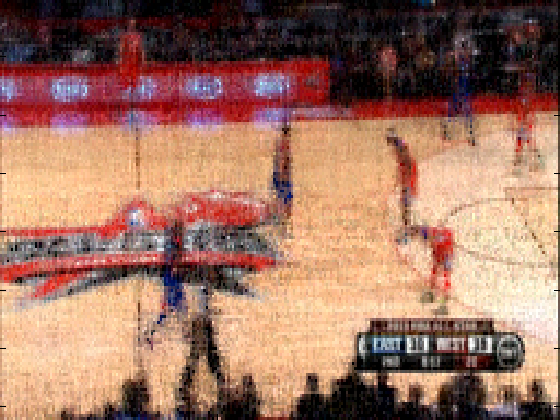} \\
      \end{tabular}}
  \caption{ Recovery for color basketball video: \textbf{Left}: Sampled Video($10\%$). \textbf{Middle}: LRTC recovery. \textbf{Right}: Tensor-nuclear-norm minimization recovery }
\label{fig:basketball_color}
\end{figure}

\begin{figure}[htbp]
\centering \makebox[0in]{
    \begin{tabular}{c c }
    \includegraphics[width = .25\textwidth, height = 1.3in]{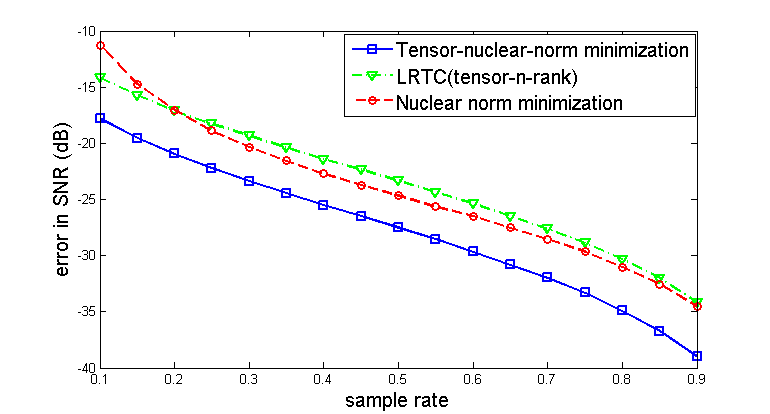} 
    \includegraphics[width = .25\textwidth, height = 1.3in]{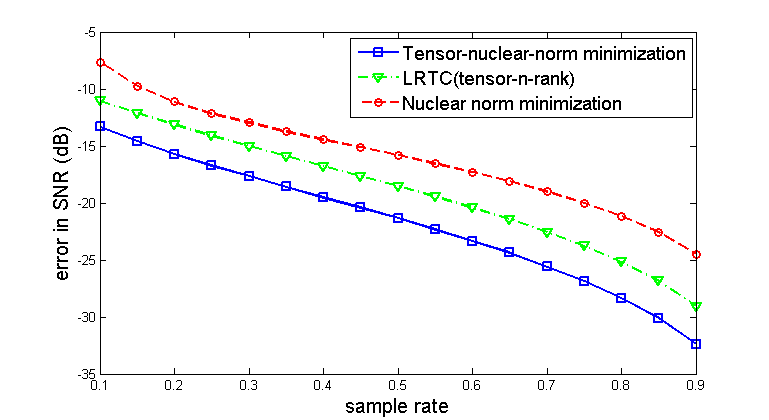}
    \end{tabular}}
  \caption{ RSE (dB) plot against sampling rate \textbf{Left}: MERL video. \textbf{Right}: Basketball video }
\label{fig:recovery_rse}
\end{figure}

Figures ~\ref{fig:recovery} and \ref{fig:basketball_color} show the results of recovery using the 3 algorithms. Figure~\ref{fig:recovery_rse} shows the RSE (dB) plots for sampling rates ranging from $10\%$ to $90\%$ where the sampling rate is defined to be the percentage of pixels which are known. Results from the figures show that the TNN minimization algorithm gives excellent reconstruction over the LRTC and Nuclear norm minimization.  These results are in line with the compressibility results in Section~\ref{sec:compression}.

\begin{figure}[htbp]
\centering \makebox[0in]{
    \begin{tabular}{c c }
      \includegraphics[width = .24\textwidth, height = 1.3in]{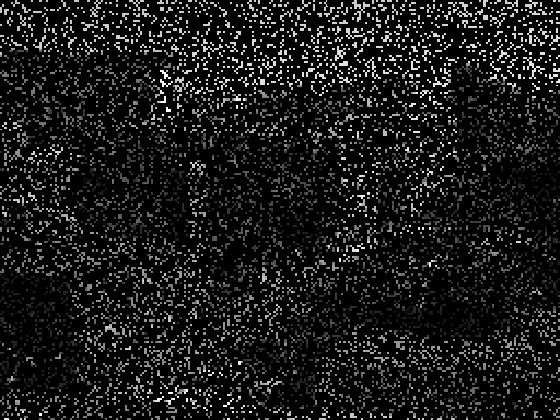}  
      \includegraphics[width = .24\textwidth, height = 1.3in]{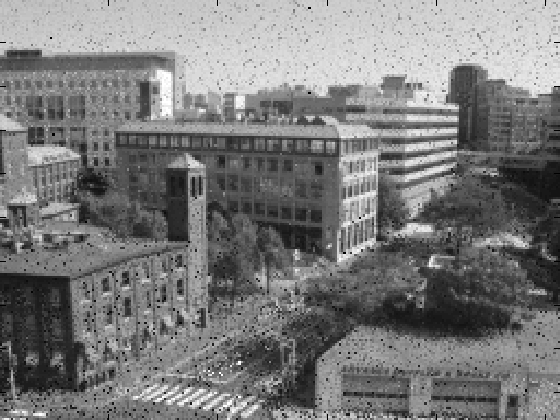}\\      
      \includegraphics[width = .24\textwidth, height = 1.3in]{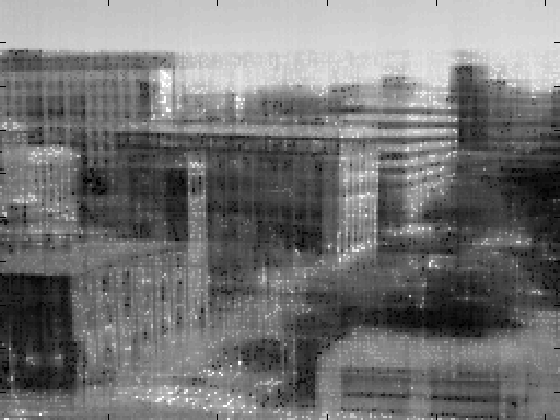}      
      \includegraphics[width = .24\textwidth, height = 1.3in]{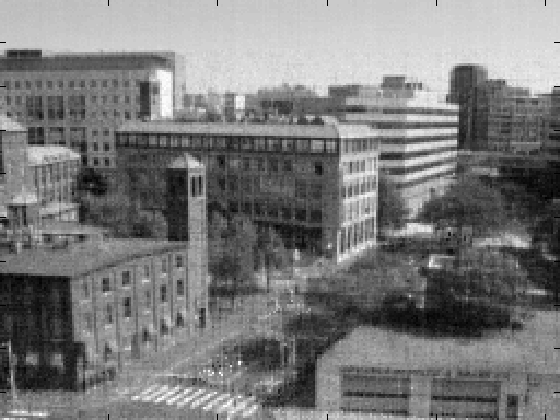}     
      \end{tabular}}
  \caption{ Tensor completion results for MERL video. \textbf{Upper left}: Sampled video($20\%$). \textbf{Upper right}: Nuclear norm minimization (vectorization and SVD based) result. \textbf{Lower left}: LRTC result. \textbf{Lower right}: TNN minimization result.}
\label{fig:recovery}
\end{figure} 

\begin{figure}[htbp]
\centering \makebox[0in]{
    \begin{tabular}{c c }   
      \includegraphics[width = .24\textwidth, height = 1.3in ]{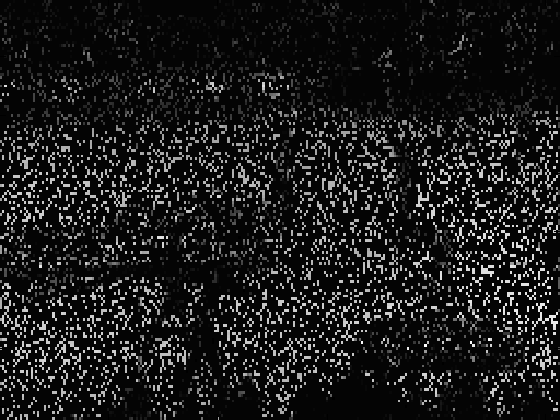}     
      \includegraphics[width = .24\textwidth, height = 1.3in]{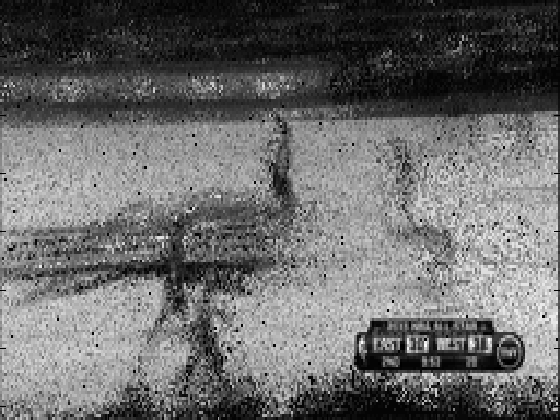}\\    
      \includegraphics[width = .24\textwidth, height = 1.3in]{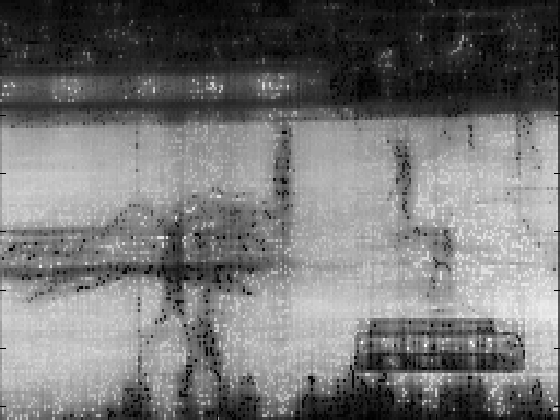}     
      \includegraphics[width = .24\textwidth, height = 1.3in]{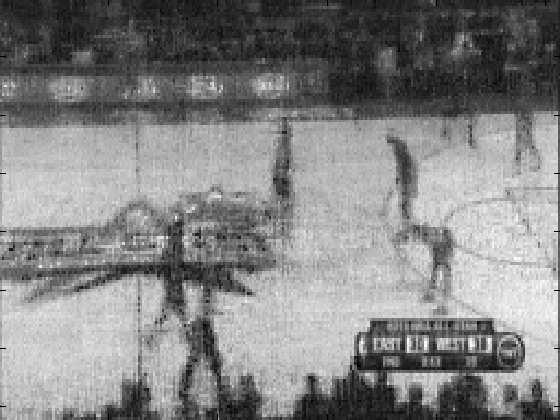} 
      \end{tabular}}
  \caption{ Tensor completion results for basketball video. \textbf{Upper left}: Sampled video($20\%$). \textbf{Upper right}: Nuclear norm minimization (vectorization and SVD based) result. \textbf{Lower left}: LRTC result. \textbf{Lower right}: TNN minimization result.}
\label{fig:recovery}
\end{figure}

%% file: rPCA.tex
\section{Tensor robust PCA}
\label{sec:rPCA}
\begin{figure}[htbp]
\centering \makebox[0in]{
    \begin{tabular}{c c}
      \includegraphics[width = .24\textwidth]{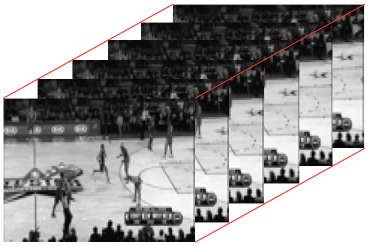}      
      \includegraphics[width = .24\textwidth]{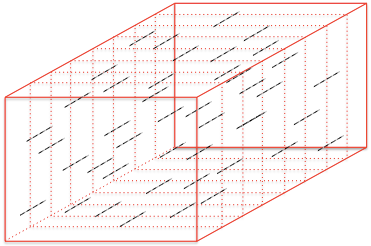}\\
      \includegraphics[width = .24\textwidth]{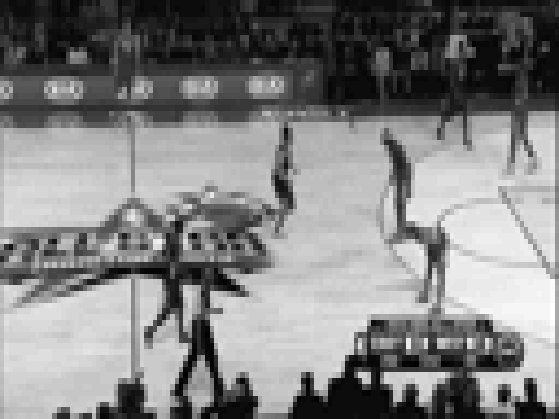}      
      \includegraphics[width = .24\textwidth]{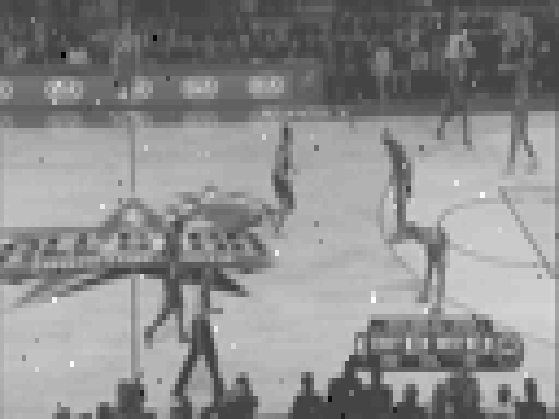} 
      \end{tabular}}
  \caption{\textbf{Upper left}: Original video. \textbf{Upper right}: Noisy tensor. For 10 consecutive frames the locations of noisy pixels are the same and then selected randomly for the next 10 frames. \textbf{Lower left} 21st frame of the original video. \textbf{Lower right} 21st frame of the noisy video.}
\label{fig:rpca1}
\end{figure}

In this section we consider a ``tensor robust principal component analysis" problem of recovering a low tensor-multi rank tensor $\T{L}$ from a sparsely corrupted observation tensor. Similar to the matrix robust PCA case \cite{Candes:2011}, suppose we have a third-order tensor $\T{M}$ such that it can be decomposed as
\begin{equation}
\T{M} = \T{L} + \T{S}
\end{equation} 
\noindent where $\T{L}$ has low tensor-multi-rank and $\T{S}$ is sparse tensor. Here we focus on a case where the sparse tensor $S$ is tubewise sparse as shown in Figure \ref{fig:rpca1}. To resolve the low rank and the sparse components given the observation $\T{M}$ we consider the following optimization problem.

\begin{equation}
\begin{aligned}
\mbox{min} \hspace{2mm}&\|\T{L}\|_{TNN} + \lambda \|\T{S}\|_{1,1,2}\\
\mbox{subject to} \hspace{2mm}&\T{M} = \T{L} + \T{S}
\end{aligned}
\label{eq:tensor_PCA}
\end{equation}

\noindent where $\lambda>0$ and the $\|\T{S}\|_{1,1,2}$ for 3-D tensors is defined as $\sum_{i,j} ||\T{S}(i,j,:)||_{F}$.

An application where this is useful arises in multilinear imaging scenarios where some pixels have heavy noise on them and the task is to automatically locate such pixels and recover the video. Although this may be done by processing each frame but if the noise artifacts and video features are aligned, one needs to both detect the noise and estimate the corrupted video feature.

In order to solve the convex optimization problem of Equation~(\ref{eq:tensor_PCA}) we use ADMM. Then we have the following recursion,
\begin{align}
\label{eq:low_rank}
&\T{L}^{k+1}=\arg\min_{\T{L}} \|\T{L}\|_{TNN} + \frac{\rho}{2} \|\T{L} + \T{S}^k - \T{M} + \T{W}^k \|^2_F\\
\label{eq:sparse}
&\T{S}^{k+1}=\arg\min_{\T{S}} \lambda \|\T{S}\|_{1,1,2} + \frac{\rho}{2} \|\T{L}^{k+1} + \T{S} - \T{M} + \T{W}^k \|^2_F\\ 
\label{eq:dual}
&\T{W}^{k+1} = \T{W}^k + \T{L}^{k+1} + \T{S}^{k+1} - \T{M}
\end{align}
\noindent where $\T{W} = \rho \T{Y}$. From section \ref{sec:tcompletion} we already have the solution to (\ref{eq:low_rank}) if we transform this equation into the Fourier domain then the tensor-nuclear-norm of $\T{L}$ will be the nuclear norm of $\text{blkdiag}(\hat{\T{L}})$. Let $\T{D}^k = \T{M} - \T{W}^k - \T{L}^{k+1}$, then the update of Equation~(\ref{eq:sparse}) is given by
\begin{equation}
\label{eq:sparse2}
\T{S}^{k+1}(i,j,:)= \left(1-\frac{\lambda}{\rho \|\T{D}^k(i,j,:)\|_F}\right)_+ \T{D}^k(i,j,:)
\end{equation}
\noindent where $i = 1,2,...,n_3$.

For experiment we consider a video, which is compressible in the t-SVD. We randomly corrupt video data by corrupting some pixels with heavy additive noise. We want to estimate the locations of such pixels using tensor robust PCA. The video used in this application is the basketball video with randomly chosen sparse pixels tubes along the third dimension. For each selected pixel we add random Gaussian noise on it. Figure~(\ref{fig:rpca1}) shows the original video(tensor) and the noise tensor.  The size of each frame is $72 \times 128$ and the total number of frames is 80. The noisy pixel tubes within every 10 frames are consistent. We use the above ADMM algorithm to separate the original video and the noise. Our analysis (to be reported in a future paper) shows that the optimal choice of $\lambda$ for tensor robust PCA is $\frac{1}{\sqrt{\text{max}(n_1,n_2)}}$. We also perform matrix robust PCA on this noisy video data by vectorizing each frame, saving it as a column vector and then get a $n_1n_2 \times n_3$ matrix. In this case the choice of $\lambda$ is $\frac{1}{\sqrt{\text{max}(n_1n_2,n_3)}}$~\cite{Candes:2011}.

\begin{figure}[htbp]
\centering \makebox[0in]{
    \begin{tabular}{c c}
      \includegraphics[width = .24\textwidth, height = 1.1in]{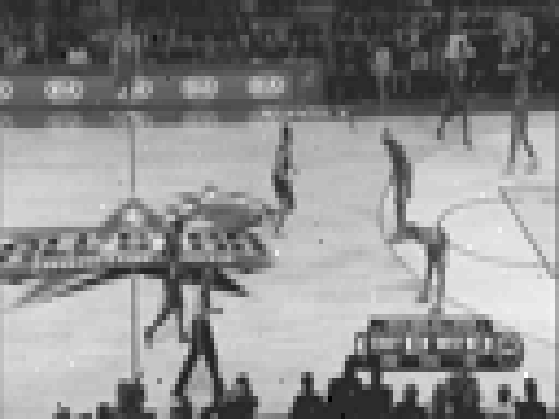}      
      \includegraphics[width = .24\textwidth, height = 1.1in]{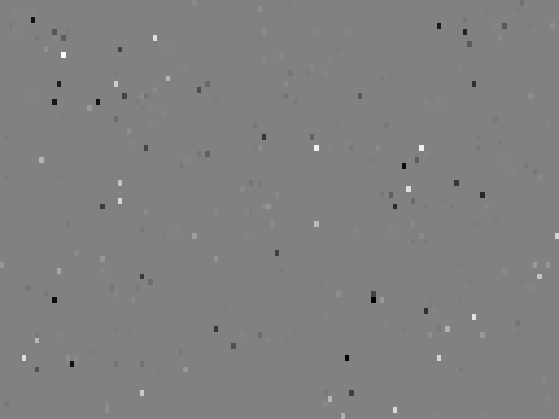} \\
      
      \includegraphics[width = .24\textwidth,height = 1.1in]{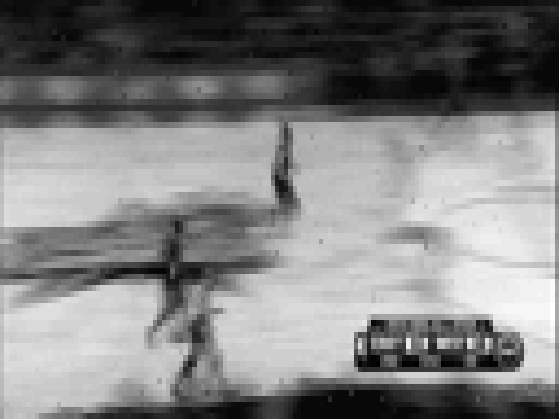}      
      \includegraphics[width = .24\textwidth,height = 1.1in]{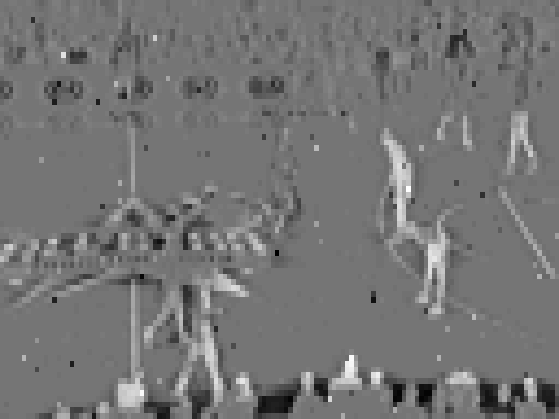}
            
      \end{tabular}}
  \caption{(21st frame shown) \textbf{Upper Left}: Low tensor multi-rank part recovered from tensor robust PCA. \textbf{Upper Right}: Sparse reconstruction from tensor robust PCA. \textbf{Lower left}: Low matrix rank part recovered from matrix robust PCA. \textbf{Lower right}: Sparse reconstruction from matrix robust PCA. }
\label{fig:rpca2}
\end{figure}

 The result of both tensor robust PCA and matrix robust PCA is shown in Figure~\ref{fig:rpca2}. From the results we can see that tensor robust PCA works very well on separating the noisy pixels from the video. However, the matrix robust PCA results in an almost fixed blurred background as the low rank part while some structure of the playground, moving people and the noise are recovered as the sparse part.

%% file: Conclusion.tex
\section{Conclusion and Future work}
\label{sec:future_work}
\vspace{-2mm} 
In this paper we presented novel methods for completion and de-noising (tensor robust PCA) of multilinear data using the recently proposed notion of tensor-SVD (t-SVD). As an application we considered the problem of video completion and de-noising from random sparse corruptions, and showed significant performance gains compared to the existing methods. The t-SVD based tensor analysis and methods can handle more general multilinear data as long as the data is shown to be compressible in the t-SVD based representation, as has been recently shown for pre-stack seismic data completion in \cite{Ely_SEG2013}. Finding the necessary and sufficient conditions for recovery of low (multi)rank tensors using TNN from incomplete tensor data is an important theoretical problem and is an important area of future research. 

\section{Acknowledgements}
This research was supported in part by the National Science Foundation grant NSF:1319653.